\documentclass{article}

\usepackage{arxiv}

\usepackage[utf8]{inputenc}
\usepackage[T1]{fontenc}
\usepackage{graphicx}

\usepackage{makecell}
\usepackage{float}

\usepackage[font={small}]{caption}
\usepackage[justification=centering]{caption}
\usepackage{soul,xcolor}
\usepackage{booktabs}
\usepackage{enumerate}
\usepackage{textgreek}
\usepackage{array}
\usepackage{multirow}
\usepackage{boldline}
\usepackage{tablefootnote}
\usepackage{amsmath}
\usepackage{xcolor}

\usepackage[hyperfootnotes=false]{hyperref}       
\usepackage{url}            
\usepackage{amsfonts}       
\usepackage{nicefrac}       
\usepackage{microtype}      
\usepackage{lipsum}		
\usepackage{doi}

\usepackage{tabularx}

\newcommand\blfootnote[1]{%
  \begingroup
  \renewcommand\thefootnote{}\footnote{#1}%
  \addtocounter{footnote}{-1}%
  \endgroup
}

\title{ 

Trimming Feature Extraction and Inference for MCU-based Edge NILM: a Systematic Approach

}

\date{\vspace{-5ex}}

\author{ 

    {Enrico~Tabanelli}\\
		Department of Electrical, Electronic\\and Information Engineering \\
		University of Bologna \\ 
		Bologna, Italy 40136 \\
		\texttt{enrico.tabanelli3@unibo.it} \\

    \And

    {Davide~Brunelli}\\
		Department of Industrial Engineering\\
		University of Trento \\ 
		Trento, Italy 38123 \\
		\texttt{davide.brunelli@unitn.it} \\

    \And

    {Andrea~Acquaviva}\\
		Department of Electrical, Electronic\\and Information Engineering \\
		University of Bologna \\ 
		Bologna, Italy 40136 \\
		\texttt{andrea.acquaviva@unibo.it} \\

    \And

    {Luca~Benini}\\
		Department of Information Technology\\and Electrical Engineering\\
		ETH Zurich\\
		Zurich, Switzerland 8092 \\
		\texttt{lbenini@iis.ee.ethz.ch} \\
}

\renewcommand{\shorttitle}{\textit{arXiv} Template}

\begin{document}

\maketitle

\begin{abstract}
Non-Intrusive Load Monitoring (NILM) enables the disaggregation of the global power consumption of multiple loads, taken from a single smart electrical meter, into appliance-level details. State-of-the-Art approaches are based on Machine Learning methods and exploit the fusion of time- and frequency-domain features from current and voltage sensors. Unfortunately, these methods are compute-demanding and memory-intensive. Therefore, running low-latency NILM on low-cost, resource-constrained MCU-based meters is currently an open challenge. This paper addresses the optimization of the feature spaces as well as the computational and storage cost reduction needed for executing State-of-the-Art (SoA) NILM algorithms on memory- and compute-limited MCUs. We compare four supervised learning techniques on different classification scenarios and characterize the overall NILM pipeline’s implementation on a MCU-based \emph{Smart Measurement Node}. Experimental results demonstrate that optimizing the feature space enables edge MCU-based NILM with 95.15\% accuracy, resulting in a small drop compared to the most-accurate feature vector deployment (96.19\%) while achieving up to 5.45$\times$ speed-up and 80.56\% storage reduction. Furthermore, we show that low-latency NILM relying only on current measurements reaches almost 80\% accuracy, allowing a major cost reduction by removing voltage sensors from the hardware design.
\end{abstract}
\keywords{Non-Intrusive Load Monitoring \and Machine Learning \and Edge Processing \and Low-Power MCUs}
%
%
\blfootnote{This work was supported by the Italian Ministry for Education, University and Research (MIUR) under the program “Dipartimenti di Eccellenza (2018-2022)”.}
\blfootnote{Moreover, this research is funded by ECSEL, the Electronic Components and Systems for European Leadership Joint Undertaking under grant agreement No 826452 (Arrowhead Tools), supported by the European Union Horizon 2020 research and innovation program and by the member states.}
\section{Introduction}
\begin{figure}[h]
    \centering
    \captionsetup{justification=centering}
    \includegraphics[width=0.5\linewidth]{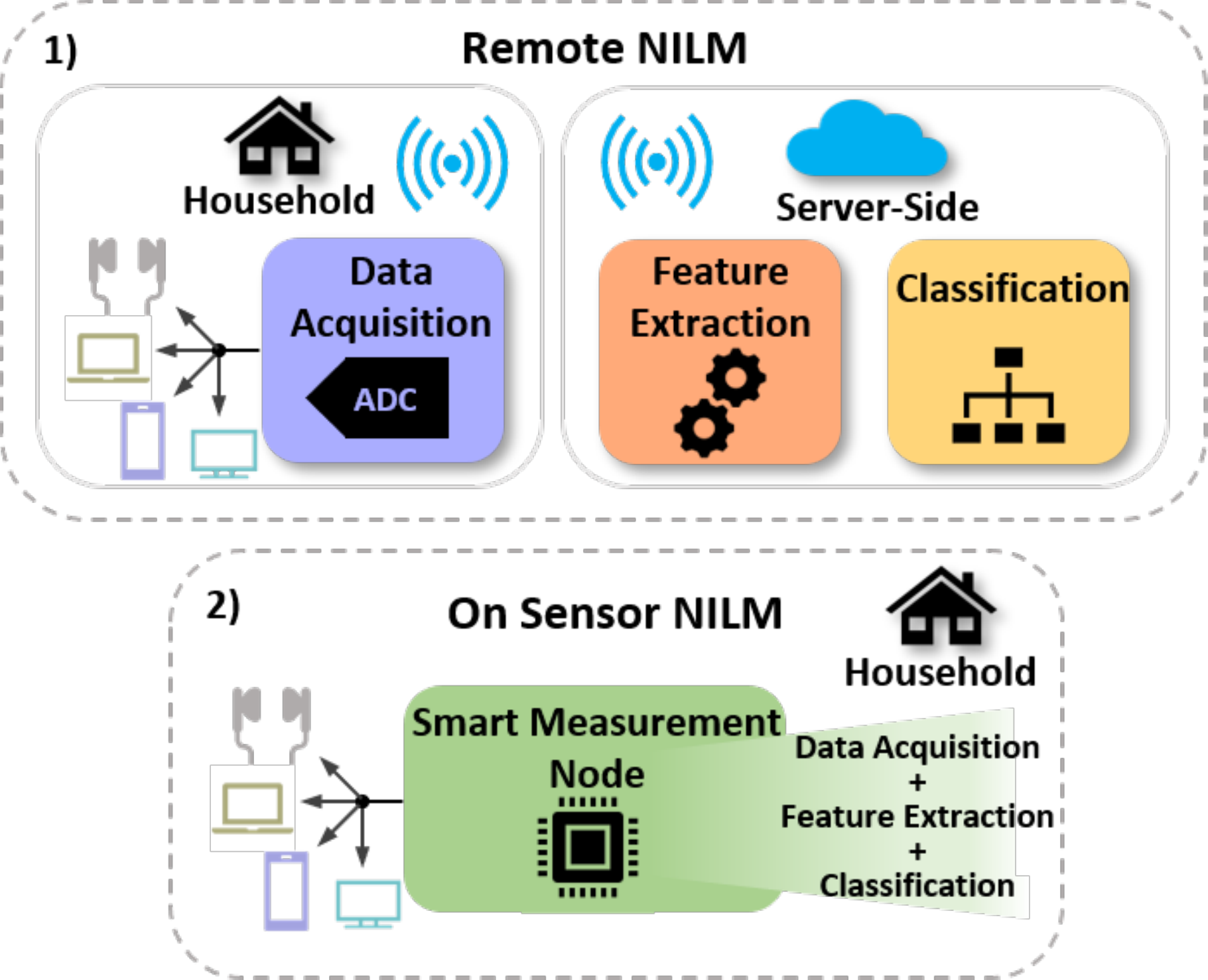}
    \caption{\textbf{Non-Intrusive Load Monitoring Architecture} \\ 1) Cloud-Based 2) Edge-Based.}
    \label{fig:NILM_Intelligence}
\end{figure}

Non-Intrusive Load Monitoring (NILM) enables the disaggregation of the electric power consumption of individual appliances from a single measurement point. Modern smart meters indeed allow reading voltage and current data almost in real-time. Coupled with NILM disaggregation, it is possible to obtain the power breakdown of energy loads without deploying distributed on-appliance metering nodes, thus increasing flexibility and reducing costs. While NILM has been studied for decades and nowadays, effective algorithms are available, it has been adopted mainly for statistical information collection on a daily basis. Today, innovative services can be potentially delivered based on near real-time edge-based load recognition, such as anomaly detection in industrial appliances and increased security in domestic contexts. \\
State-of-art NILM approaches leverage high-dimensional feature spaces and computational resource-demanding ML algorithms to bring tangible benefits in load disaggregation accuracy~\cite{c1}. Indeed, multi-feature approaches impose high memory and computing requirements making cloud-computing deployment mandatory~\cite{c2}. Figure~\ref{fig:NILM_Intelligence} shows the standard server-based NILM framework architecture in a household implementation. A local meter performs power measurements, while a 
server-side back-end performs compute-intensive feature extraction and classification algorithms. High bandwidth communication is required for data uplink between the two sides.
\\
Cloud-computing frameworks suffer from scalability issues in terms of communication latency, bandwidth, and privacy~\cite{c3}. Furthermore, they could put at risk customers privacy, revealing energy profiles, and daily activities~\cite{c4}. Bringing NILM execution on-edge (or even on the smart meter board) would enable faster end-user corrective actions to reduce energy consumption, react to damages, or trigger alarms. For instance, recent studies highlight how near real-time (i.e., during monitoring) feedback on appliance power consumption could lead to energy savings over 5-20\%, against 0-10\% of cloud-based systems~\cite{c5}.
\\
The key practical challenge with edge-NILM is that low-cost MCU-based devices (either edge or metering nodes) have very limited resource budgets, especially SRAM memory and Flash storage. The on-chip memory is 6 orders of magnitude smaller than cloud-based systems, making unmodified cloud algorithms deployment on MCU unfeasible. Consequently, there is need to explore the memory-latency-accuracy trade-off, leveraging optimized feature spaces to achieve memory-efficient and lightweight frameworks suitable for edge devices.\\
An additional challenge for edge-NILM is that many SoA algorithms use features requiring high load sampling frequency. Low-cost commercial smart meters lack proper analog front-ends, capable of a suitable sampling frequency, as well as digital back-ends for processing of the high-throughput digitized samples. Research-oriented solutions use high-end platforms not suitable for low-power and low-cost edge deployment. To overcome both these shortcomings, we exploit a custom MCU-based \emph{Smart Measurement Node} described in~\cite{c6}. The meter features an adequate analog back-end, allowing extracting a wide range of features moving beyond standard low-frequency features extracted by commercial meters. On this device, we developed a novel approach to enable the deployment at the edge of four SoA Machine Learning NILM algorithms on different load scenarios. We explore the memory-latency-accuracy trade-off by varying feature dimensionality, and we pinpoint optimal characterization points leading to lightweight models without sacrificing model performance. The methodology proposed allows moving from expensive high-end platforms toward more cost-efficient edge solutions.\\
The contributions of this paper are:
\begin{enumerate}
    \item We designed a NILM framework, consisting of data extraction and classification software modules optimized for edge devices such as the \emph{Smart Measurement Node}. To this purpose, we performed a Mean Decrease Accuracy analysis to reduce the feature space with minimal information loss. We thus identified the most relevant time- and frequency-domain features in disaggregating load profiles depending on the classification scenarios.
    \item Using the developed framework, we present a comprehensive investigation of memory size, performance, and accuracy in feature extraction and classification stages, characterizing SRAM/Flash memory requirements and execution cycles. According to worst-case extraction requirements (full feature vector), we determine classification run-time constraints: 8.295~Mcycles, 498~kB Flash storage and 72~kB SRAM memory. 
    \item We compare four memory-constrained supervised learning techniques available in the literature on different load classification scenarios, highlighting their trade-offs in terms of memory size, performance, and accuracy. 
    \item We reduce the feature space to 5 components enabling a RF-based edge-NILM with 95.15\% accuracy. The optimized feature vector results in a small accuracy drop (1.04\%) compared to the most-accurate feature vector deployment (96.19\%) while reaching up to 5.45$\times$ speed-up and 80.56\% storage reduction.
    \item We demonstrate that adding frequency-domain features enhances accuracy by almost 3\% with respect to using only time-domain components (86.66\%), thus attaining 89.84\%. Furthermore, we show that 35 frequency-domain features enable a low-latency edge-NILM to reach roughly 80\% accuracy, with nearly no drop compared to full feature deployment, while leading to a significant cost decrease in removing voltage sensors from the hardware design.
\end{enumerate}
The paper is organized as follows. In Section II, we present related works, and we discuss the latest approaches for load disaggregation. Section III describes the hardware of the \emph{Smart Measurement Node}. Section IV highlights the crucial stages of the NILM framework, including data acquisition, features extraction, and disaggregation. Experimental results of our exploration of feature extraction and disaggregation algorithms are reported in section V. Concluding remarks and plans for future works complete this paper.
\section{Related Work}
Since Hart et al.~\cite{c7} firstly introduced NILM, several works proposed steady-state model-driven approaches to distinguish appliances based on different feature spaces. The method performs well on ordinary ON/OFF appliances but fails when applied to more sophisticated ones, such as Finite State Machine (FSM) loads or Continuously Variable Devices (CVD). On the other hand, also using transient-state features~\cite{c8} can offer useful appliance-level information. However, a high sampling rate is mandatory to collect reliable measurements, and different turn-on and off transients raise significant issues.\\
Along with the growing interest in ML, ML-based NILM methods gained popularity. J. Kelly et al.~\cite{c9} firstly applied Neural Networks (NN) to NILM using data sampled at 1Hz. A recent approach~\cite{c10} extracts EMI features in the frequency domain sampling at 1MHz and classifies using a kNN algorithm. To handle the issue of labeling a massive amount of data, in~\cite{c11}, the authors present a semi-supervised multi-label Temporal Convolutional Network (TCN)-based framework to extract load signatures from the aggregate real power (P). While Basu et al.~\cite{c12} focused on generating meta-features from energy readings to improve disaggregation performance, Bernard et al.~\cite{c1} proposed the fusion of low, mid, and high-frequency features. These works exhibit outstanding accuracy in monitoring several types of loads. However, the high computational and memory demand prevents running real-time NILM based on these methods on low-cost MCU-based devices.\\
To avoid the computational limitation of MCUs, many solutions rely on cloud-based systems. In~\cite{c13}, the authors acquire measurements from local meters with a sampling rate of up to 3kHz, while a cloud framework runs fully-connected NNs. Green et al.~\cite{c14} leverage a powerful cloud back-end which processes multi-feature vectors with multiple algorithms and combines their output to enhance load recognition.\\
Recently the design of online power meters has gained an increasing interest in several research groups. Barsocchi et al.~\cite{c15} proposed an easy-to-install led-probe-based smart meter to collect low-frequency power features and a Finite State Machine (FSM) running on the integrated Arduino platform. On the market side, Neurio and Open Energy developed low-cost easy-to-install Raspberry- and Arduino-based commercial devices. The meters support low sampling rates enabling low-frequency features acquisition for monitoring household power consumption. Unfortunately, none of the previous solutions features high-frequency analog front-ends and proper onboard resources to perform advanced on-sensor processing. Their deployment demands transmitting data to third-party cloud services to retrieve appliance-level information. \\
A few works focused on designing sensing devices featuring high-end platforms. In~\cite{c16}, the authors deploy a DAQ card coupled with a high-cost high-power E660 Intel Atom Processor to extract power features and run an ANN. Sirojan et al.~\cite{c17} proposed a MLP technique developing a meter by use of a Xilinx Field-Programmable Gate Array (FPGA) and ARM Cortex-A9 Processor, both integrated on the National Instruments (NI) myRIO-1900. These smart meters are too costly and power-demanding for commercial edge solutions.

\section{System Architecture}
\begin{figure}[h]
    \centering
    \captionsetup{justification=centering}
    \includegraphics[width=0.6\linewidth]{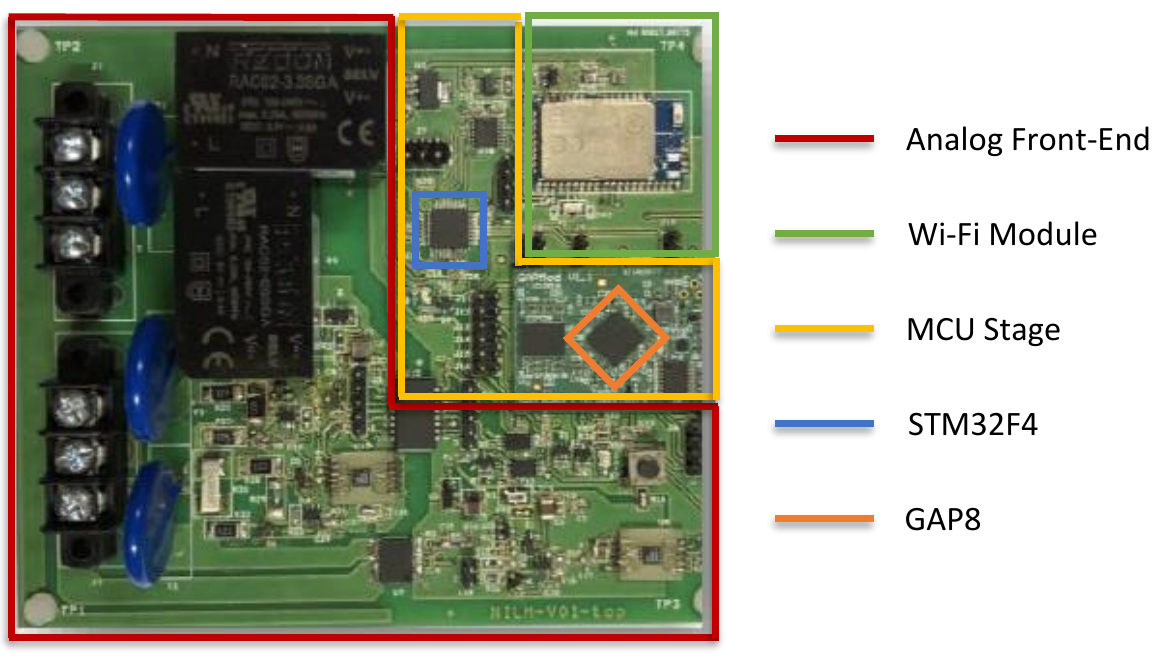}
    \caption{\textbf{Smart Measurement Node}}
    \label{fig:Schematic}
\end{figure}

This section describes the hardware platform designed and deployed for the experiments: the \emph{Smart Measurement Node}. The meter integrates two microcontroller units (MCUs), as shown in Figure \ref{fig:Schematic}. The STM32F4 is a high-performance 32-bit STMicroelectronics (STM) MCU based on the ARM Cortex-M4 core with 512kB of Flash memory and 96kB of SRAM. Running at 84MHz, the CPU delivers 105 DMIPS/285 CoreMark performance executing from Flash memory, with 0-wait states due to STM’s Adaptive Real-Time (ART) accelerator, which speeds up instruction fetch accesses to on-chip memories. The dynamic power scaling enables the current consumption to be as low as 128$\mu$A/MHz in run mode, while 9$\mu$A in stop mode. The Cortex-M4 implements an extension of the Thumb/Thumb-2 Instruction Set Architecture (ISA) supporting DSP instructions, such as single-cycle 16/32-bit, single-cycle dual 16-bit MAC, 8/16-bit SIMD arithmetic, but also saturation arithmetic and HW divide. The presence of the single-precision Floating Point Unit (FPU) improves a wide range of addressable applications. The second MCU is GAP8, a commercial 32-bit ultra-low-power IoT-edge computing engine that embeds a RISC-V multi-core processor derived from the PULP open-source project. In this work, the STM32F4 is in charge of measurement settings, data acquisition and processing, and eventually streaming results to a server. In future works, we will deploy GAP8 to speedup and parallelize NILM algorithms at the edge. \\
To acquire voltage and current samples, the Smart Measurement Node includes an analog front-end consisting of two LTC1407A modules, a dual-channel Analog-to-Digital Converter (ADC) from Linear Technology. The ADC features a sampling rate of 1.5~Msps while recording simultaneously and a 14~bit resolution with 16384 discrete digital values. Consequently, the 0-2.5~V unipolar full-scale input range results in a voltage resolution of \(152~\mu \)V. The 80~dB Common-Mode Rejection Ratio (CMRR) at 100kHz enables to remove common-mode noise properly by measuring signals differentially from the source. Moreover, the 74dB Signal-to-Noise Ratio (SNR) at 100kHz enhances the ADC low-noise performance, while the 14mW power dissipation emphasizes its energy efficiency. In addition, the analog stage offers an Isolated Interface, consisting of a voltage divider and a Shunt resistor to measure voltage and current, and a Non-Isolated Interface, enabling the usage of Rogowski Coils and Hall-Effect Sensors. We deployed the Isolated Interface in our work because it enables direct and simultaneous current and voltage sampling.\\
The board additionally embeds the WF121 Wi-Fi module by Bluegiga Technologies. The device provides a 2.4~GHz 802.11~b/g/n radio and a 32-bit MCU, which offers low-level programming drivers and an API for several applications. Sending 256-byte-sized packets to a server awaiting the receiving end, we tested the Wi-Fi bandwidth. The test resulted in a bandwidth of 800~kbps on a 2.56~Mb transmission size, which translates into an upload sample rate of 57~ksps (28.5~ksps, respectively) with a 14-bit sample resolution.\\
The firmware starts with a preparatory phase, which sets up a Wi-Fi connection to the client-server and puts the Wi-Fi module in streaming mode. By setting with an advanced control timer a sampling rate of 20kHz, the ADC starts collecting current and voltage measurements. Since STM HAL low-level drivers support only multiples of bytes, we store two 14-bit samples in 4-Byte-Arrays, and we mark buffer overflows using the 4 remaining bits. The Serial Peripheral Interface (SPI) streams acquired data operating at a frequency of 16Mbps, while on the active MCU, the Direct Memory Access (DMA) manages the reception. We average two successive measurements to reduce the noise, which ends in an effective sampling rate of 10kHz. An interrupt triggers the MCU, which extracts the features and starts the classification process. We transmit the result via Universal Synchronous-Asynchronous Receiver/Transmitter (USART) to the Wi-Fi module, therefore readily streamed to the receiving client. 

\section{NILM framework}
\begin{figure}[h]
    \centering
    \captionsetup{justification=centering}
    \includegraphics[width=0.8\linewidth]{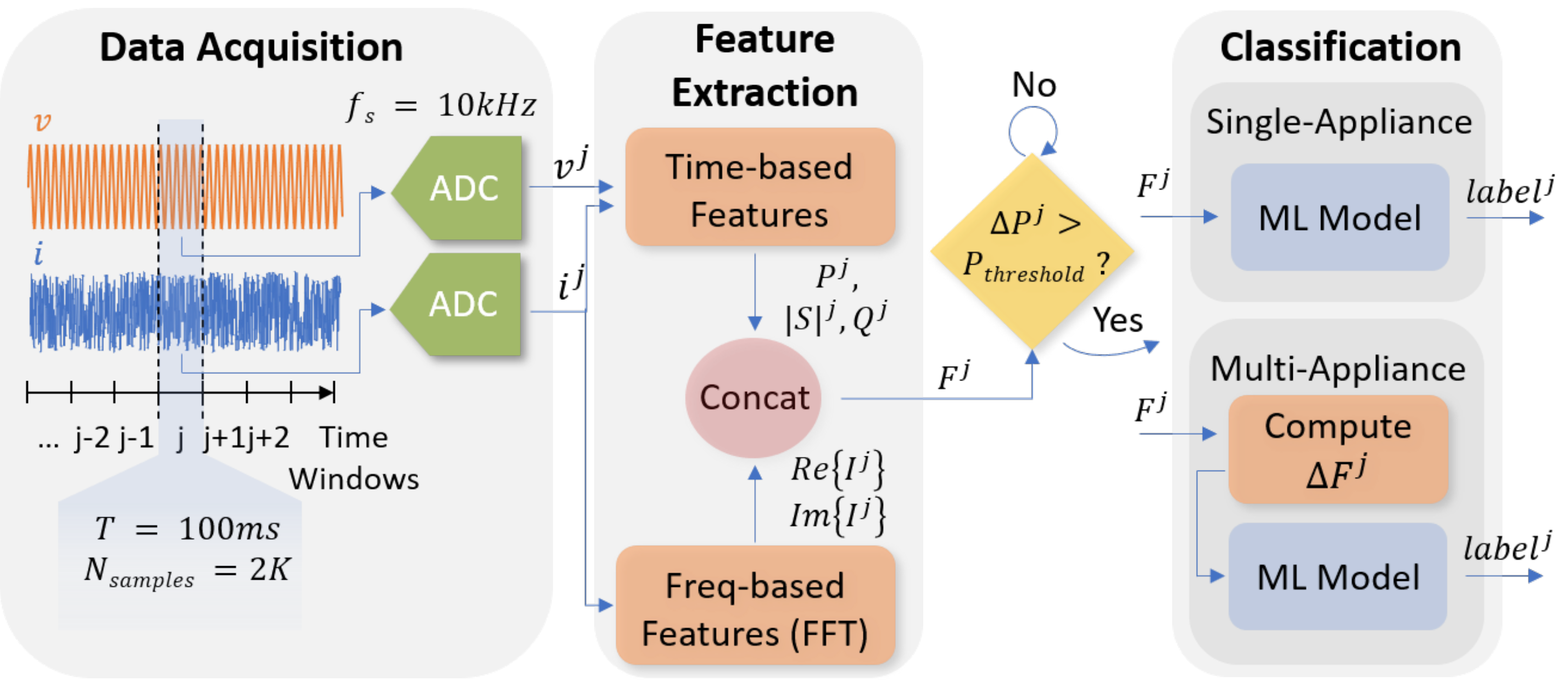}
    \caption{\textbf{NILM Framework}}    
    \label{fig:NILM_framework}
\end{figure}

\begin{figure}[h]
    \centering
    \captionsetup{justification=centering}
    \includegraphics[width=0.7\linewidth]{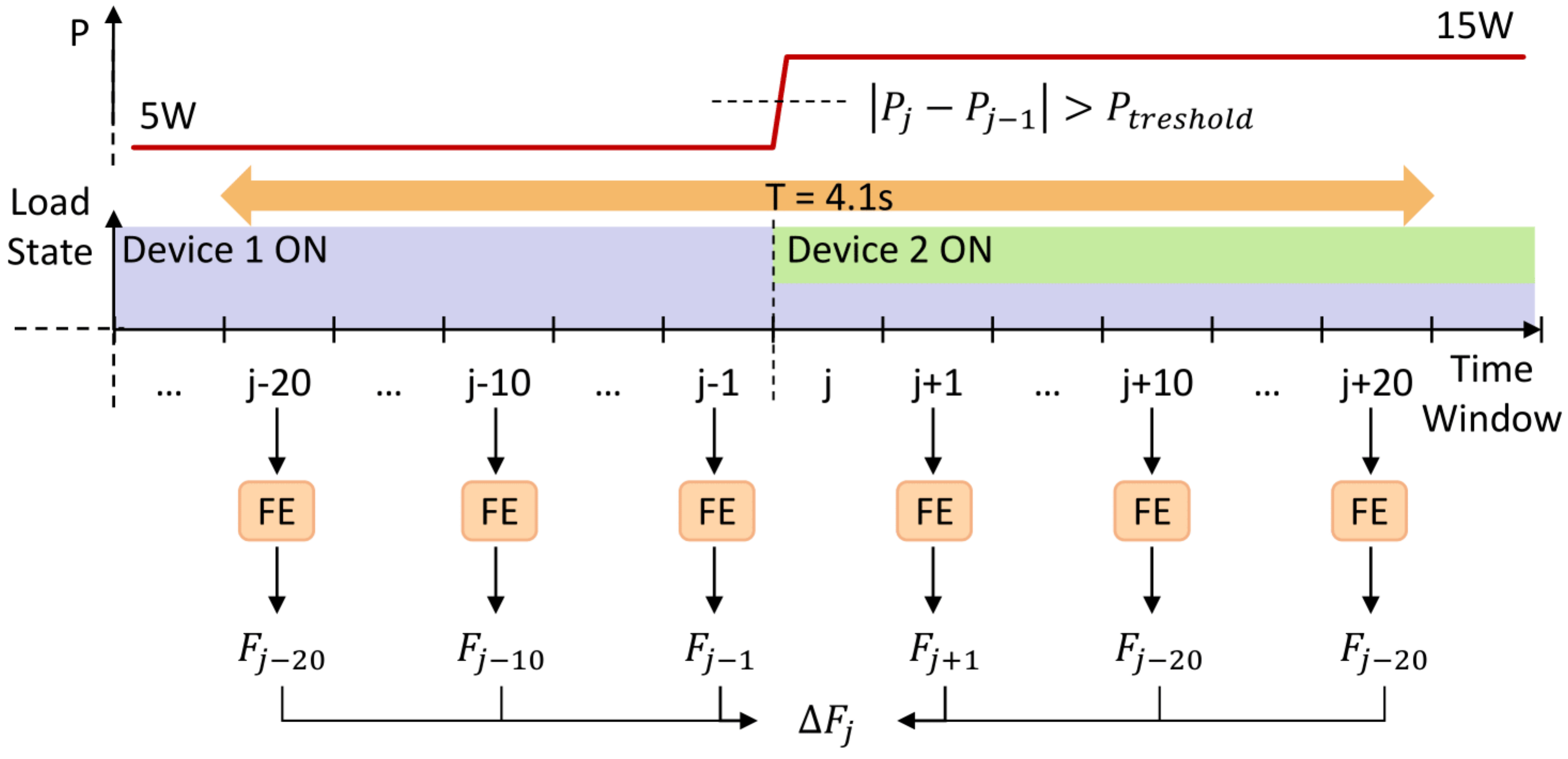}
    \caption{\textbf{$\Delta F$ Calculation}}
    \label{fig:Delta}
\end{figure}

As depicted in Figure \ref{fig:NILM_framework}, our NILM framework consists of three stages. A {\em data acquisition} system collects voltage and current measurements at a sampling frequency of 10kHz. When a 100ms time-window of acquisitions (1000 samples per channel) is available, the {\em feature extraction} stage extracts time- and frequency-domain features. Thus, if a variation of $P$ exceeding a pre-characterized threshold is detected (i.e., a switching event), the framework enters the disaggregation state in which the classification is performed. The disaggregation methodology differs according to the classification scenario:

\begin{itemize}

    \item {\em Single-Appliance}: The recognition method works in a simplified setting with only one active appliance at a time. When $P$ exceeds the threshold in the time-window $j$, we feed the extracted feature vector to a ML model, which attempts to recognize the active device providing a label for $j$. This setting can be considered as a reference for the accuracy results in the more complex multi-appliance scenario. Also, we implemented a baseline scenario ({\em Single-Appliance \#1}) and a more challenging one ({\em Single-Appliance \#2}) to study the impact of using frequency domain features when loads have similar time-domain features. Using a single appliance at a time allowed to better understand the impact of these features on load recognition.
    
    \item { \em Multi-Appliance}: This recognition method supports the case in which multiple appliances are active in overlapped time windows. It works by observing features in a time interval across a switching event. For this reason, a constraint for this method to work is that two switching events from different appliances do not take place during the observation interval. When a variation of $P$ over the threshold is detected, we calculate the differential feature vector $\Delta F$, described in Equation \eqref{eq:1}. As depicted by Figure~\ref{fig:Delta}, $\Delta F$ combines features from different intervals around the switching event marked by the $P$ variation. Precisely, we consider features in the time windows immediately preceding ($F_{j-1}$) and following ($F_{j+1}$) the event, as well as 10 and 20 time windows before ($F_{j-10}$, $F_{j-20}$), and after ($F_{j+10}$, $F_{j+20}$) the event. The averaged feature vectors result in two intermediate vectors, whose subtraction leads to $\Delta F$. Thus, the ML model tries to infer the activated load returning a class for the time-window $j$. To correctly compute the $\Delta F$, it is assumed that loads are not turned on simultaneously within the overall $\Delta F$ windows, that is $41*100ms=4.1s$. As discussed in Section~\ref{sec:disaggregation}, this method achieves slightly lower accuracy compared to Single-Appliance methods but enables multiple load recognition.  

\end{itemize}

\begin{equation}
    \Delta F_j=\frac{F_{j-20} + F_{j-10} +F_{j-1}}{3} - \frac{F_{j+1} + F_{j+10} + F_{j+20}}{3}\label{eq:1} 
\end{equation}

Concerning the power threshold, our previous work~\cite{c6} includes a characterization study of the power threshold, which led to 5W as the most effective one for detecting switching events.

\subsection{Data Acquisition}
The data acquisition system acquires aggregated load measurements at a sampling frequency of 10kHz to identify distinctive load patterns in both Single-Appliance and Multi-Appliance scenarios. To train the disaggregation algorithms, we adopted the Domestic Appliances Dataset (DAD) collected with the \emph{Smart Measurement Node} in~\cite{c6}, openly accessible at~\cite{c18}. The recorded appliances are both linear and non-linear and belong to the following categories described in the literature~\cite{c7}: ON/OFF loads, Finite State Machine (FSM) appliances, and Continuously Variable Devices (CVD). We divided the DAD into 3 sections to set up 2 Single-Appliance and 1 Multi-Appliance datasets, described in Section~\ref{sec:disaggregation}.
\par

\subsection{Feature Extraction}
To identify active appliances, a set of high-quality features must be extracted from raw measurement data. Since NILM features are highly dependent on the sampling rate, we divide them between time- and frequency-domain features in this work. Among the plethora of possible choices, we picked features that already proved their ability to enhance the disaggregation process. As claimed by \cite{c9}, the use of power-related features allows discriminating simple linear loads. For that purpose, within the 100ms-long time-frame, we calculate the instantaneous power, then we average it and compute the Real ($P$), Reactive ($Q$), and Apparent Power ($|S|$). However, these simple time-domain features lack effectiveness with FSM and CVD loads. In \cite{c10} the authors showed that Electromagnetic Interference (EMI) signals enable a finer differentiation of similar switching mode power supplies. Therefore, we compute the Fast Fourier Transform (FFT) of electric current samples over the same time-frame. The resulting sampling frequency of 10kHz enables determining current harmonics until 5kHz at a resolution of 10Hz. Since odd current harmonics represent typical features for load disaggregation~\cite{c19}, we extract 50Hz odd current harmonics. 
\par

\subsection{Mean Decrease Accuracy Analysis}
To reduce NILM computing and memory requirements, we performed a Mean Decrease Accuracy (MDA) Analysis~\cite{c20}. We also tested Principal Components Analysis (PCA) and t-Distributed Stochastic Neighbor Embedding (t-SNE). We report below the dimensionality reduction method giving the best results on average across all three scenarios, namely MDA analysis. The study aims at reducing the dimensionality without jeopardizing information loss. The core idea is to measure the features importance by observing the accuracy decrease when left out. After training the ML model on the full feature vector, we define the testing set accuracy as a baseline. To increase robustness in feature selection, we introduce controlled noise by randomly shuffling feature values and compute the testing set accuracy with the resulting dataset. By comparing baseline vs. actual accuracy, we calculate the performance loss due to the shuffled variable. To a higher accuracy loss corresponds a more important feature.
\par

\subsection{Disagreggation Algorithms}
Edge nodes have tight constraints. For that purpose, enabling load monitoring on such platforms requires lightweight and memory-efficient models. In this work, we deployed four SoA disaggregation algorithms developed and demonstrated in cloud-based environments. In table \ref{tab:Complexity_Algorithm}, we show how the feature dimensionality ($f_{dim}$) affects their time and space complexity.\\
\emph{1) k-Nearest Neighbor (kNN)} recently gained prominence as a load monitoring classification algorithm on server-based systems~\cite{c21}. kNN non-parametric nature enables the learning of predictive functions directly from data. However, computing and storage requirements are linearly dependent on the feature space dimensionality ($f_{dim}$). \\
\emph{2) Support Vector Machine (SVM)} is a model that has proved successful in several classification scenarios~\cite{c22}. SVM rationale is to separate the feature space by finding a set of hyper-planes in high-dimensional space. Separating data with low dimensional feature spaces requires a large Support Vector (SV) set, increasing memory and computing effort. On the other hand, high dimensional feature spaces handily solve the separation problem with fewer SVs. However, high dimensionality becomes again demanding in terms of memory and computation. \\
\emph{3) Neural Networks (NN)} demonstrated a huge potential when applied to energy disaggregation~\cite{c23}. The capability to learn non-linear functions makes MLP an attractive solution for NILM. However, increasing the scenario complexity and feature space dimension, MLP becomes highly compute-demanding and memory-hungry if not properly tuned. \\
\emph{4) Random Forest (RF)} classifiers applied to load monitoring can achieve excellent results in different classification contexts~\cite{c24}. The model consists of several decision trees created at training time, which provide a class prediction for an input object. Then the model aggregates the votes to decide the final class.  The RF storage usage is highly dependent on the number of trees ($N_{trees}$) in the forest and the space dimensionality ($f_{dim}$). Thus, a methodology is mandatory to obtain a feasible model for edge-devices.
\par

\begin{table}[h]
    \centering
    \setlength\extrarowheight{2pt}
    \begin{tabular}{l|c|c}
    
        \hlineB{3}
        Model   & \textbf{Time Complexity}                                                          & \textbf{Space Complexity} \\ 
       
        \hlineB{3}
        KNN     & \emph{O}(f\textsubscript{dim} $\times$ Dataset\textsubscript{size})               & \emph{O}(f\textsubscript{dim} $\times$ Dataset\textsubscript{size}) \\
        
        MLP     & \emph{O}($\sum_{i=0}^{l-1}$ In\textsubscript{i} $\times$ Out\textsubscript{i})    & \emph{O}(N\textsubscript{weights} + N\textsubscript{bias}) \\      
   
        RF      & \emph{O}(f\textsubscript{dim} $\times$ N\textsubscript{trees})                    & \emph{O}(f\textsubscript{dim} $\times$ N\textsubscript{trees}) \\    

        SVM     & \emph{O}(f\textsubscript{dim} $\times$ N\textsubscript{sv})                       & \emph{O}(f\textsubscript{dim} $\times$ N\textsubscript{sv})  \\ 
        
        \hlineB{3}
        
    \end{tabular}
    \vspace{0.2cm}
    \caption{\textbf{Algorithms Time and Space Complexity}}    
    \label{tab:Complexity_Algorithm}
\end{table}

\section{Evaluation}
This section discusses the results of our feature space optimization strategy. Firstly, we show the computing and memory effort required to extract time- and frequency-domain features on the ARM Cortex-M4 core. Then, we introduce three load monitoring scenarios, and we describe the most significant memory-performance-accuracy trade-offs, also accompanied by precision and recall measurements. Then we compare framework run-time characteristics to determine the best-suited algorithms for edge-NILM. \\
The methodology used for each scenario and algorithm consists of the following stages:

\begin{enumerate}
    \item Initial Grid Search on the full feature vector.
    \item Mean Decrease Accuracy (MDA) analysis for sorting the features in descending order of importance. 
    \item Model training and testing by adding one feature at a time from the ranked vector. We applied a Grid Search to each point for hyper-parameters fine-tuning.
    \item Selection of feature vector points satisfying edge constraints while remaining within a 5\% accuracy drop from the most-accurate point. 
\end{enumerate}

\subsection{Feature Extraction Requirements}

Table \ref{tab:Complexity} shows the feature extraction memory occupation and execution cycles required to compute each component on the ARM Cortex-M4 core. We reported the computational effort in terms of Multiply-Accumulate (MAC) operations. Since MCU-based devices have tight memory constraints, we also report the run-time SRAM memory and Flash storage requirements. \\
\textit{RawConv} refers to the calibration procedure applied to raw ADC samples after the data logging. Multiplying and adding by gain and offset coefficients, we calculate calibrated current and voltage measurements. Converting both samples signals require 15~Kcycles. When processing only electric current samples, the MCU takes only 6~Kcycles. \\
To compute the features, we operate on a 100ms-long time-frame over 1000 samples per channel. Real ($P$) and apparent ($|S|$) power can be extracted with 17K and 11K cycles, respectively. Instead, since the reactive power ($Q$) depends directly on $P$ and $|S|$ ($Q=\sqrt{|S|^2-P^2}$), its computing effort can vary from a 79~cycles best-case to a 28~Kcycles worst-case. \\
To extract 50Hz odd current harmonics, we use the FFT routine from ARM CMSIS-DSP software library. The FFT spectrum results in 100 real and imaginary components per time-frame at the expense of almost 66~Kcycles and 18~kB of Flash memory to store twiddle coefficients and bit reversal lookup tables. Without reordering FFT components, we can save few kBs of storage and 4~Kcycles (resulting in 62~Kcycles).\\
As shown in the table, the full feature vector extraction requires almost 105~Kcycles (62 + 28 + 15~Kcycles) and 14~kB Flash storage in the worst case. To evaluate the time available for classification, we consider that extraction and classification must fit within the 100ms of the time window. Since STM32F4 operates at 84~MHz (100ms at 84MHz = 8.4~Mcycles), considering the 105~Kcycles used for the feature extraction, we have 8.295~Mcycles available for the classification task. Considering memory constraints, we have a total of 512~kB Flash memory, of which 14~kB are occupied by the extraction task (i.e., 498~kB available). Finally, the total SRAM is 96~kB, of which the extraction process takes 24~kB.

\begin{table}[h]
    \centering
    \setlength\extrarowheight{1.5pt}
    \begin{tabular}{ l|c|c|c|c}
        \hlineB{3}
                                                & \textbf{SRAM} (kB)    & \textbf{Flash} (kB)   & \textbf{MAC (k)}       & \textbf{Cycles (K)}  \\ 
        \hlineB{3}
        RawConv\textsubscript{V\&I}         &  8                    & -                     & 4      & 15     \\
        RawConv\textsubscript{I}            &  4                    & -                     & 2      & 9      \\        
        P                                   &  4                    & -                     & 2      & 17     \\
        |S|                                 &  -                    & -                     & 2      & 11     \\
        Q$^1$                               &  -                    & -                     & 0.04      & 0.08        \\
        Q$^2$                               &  4                    & -                     & 2.04      & 17     \\
        Q$^3$                               &  -                    & -                     & 2.04      & 11     \\
        Q$^4$                               &  4                    & -                     & 4.04      & 28     \\
        FFT\textsubscript{1024}             &  4                    & 17.6                  & 10.24     & 66     \\
        FFT\textsubscript{1024}$^5$         &  4                    & 14.1                  & 10.24     & 62     \\
        \hline
        Full Vector                         & 24                    & 14.1                  & 18.24     & 105    \\
        \hlineB{3}
    \end{tabular}
    \vspace{0.2cm}
    \caption{
        \textbf{Feature Extraction Resource Usage}\\
        $^1$with P \& |S|, $^2$without P, $^3$without |S|, $^4$without P \& |S|,\\$^5$without reordering FFT components. ("-" implies negligible values)
        }
    \label{tab:Complexity}
\end{table}
\par

\subsection{Disaggregation Algorithms Trade-off Analysis}
\label{sec:disaggregation}
This section reports significant trade-off analysis results on the disaggregation algorithms applied to various monitoring scenarios. For single appliance scenarios, we discuss RF and SVM models. In contrast, we discuss the MLP trade-off for the multi-appliance scenario because it is more relevant and suitable to scale for a large number of appliances. A full report of deployment results is given in Section~\ref{sec:overall}.
\begin{figure}[h]
    \centering
    \includegraphics[width=0.5\linewidth]{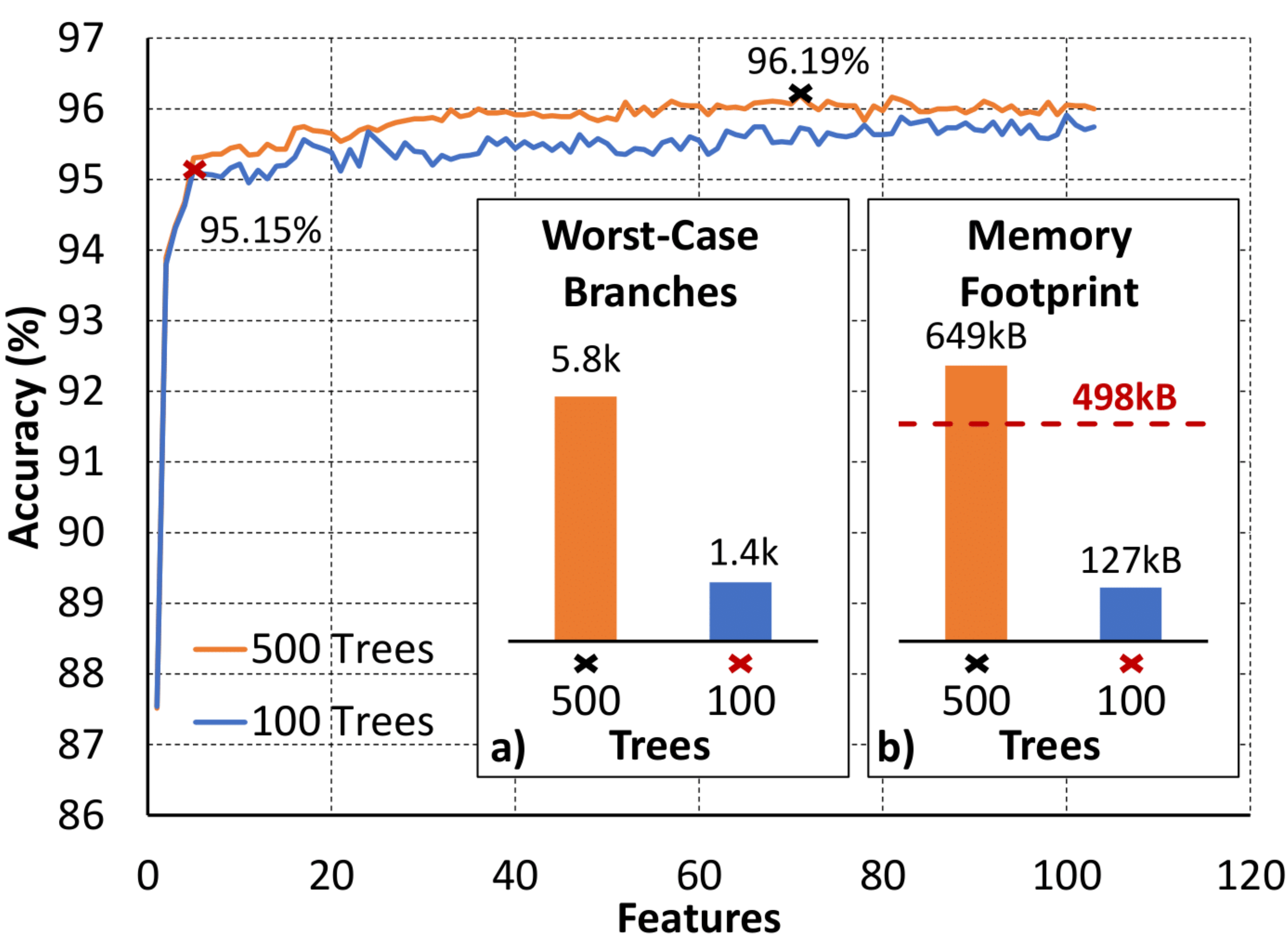}
    \caption{\textbf{RF Accuracy-Flash Trade-Off on Single-Appliance \#1 Dataset}}
    \label{fig:RF}
\end{figure}
\subsubsection{RF in Single-Appliance Scenario \#1}
The scenario investigated below represents a domestic context consisting of household appliances: cell phone charger, monitor, fan at minimum, medium and maximum speed, light bulb, and 2011 MacBook Pro in idle state. The deployed dataset comprises non-overlapping recordings resulting in 7 classes. The main plot in Figure \ref{fig:RF} represents the testing accuracy achieved using 2 RF models featuring a different number of trees (100 vs. 500). Accuracy trends up to 5 features do not change significantly. When using more than 5 features, to achieve slightly higher accuracy is necessary to raise the number of trees to 500. However, it requires a Flash storage ranging from 553kB to 727kB to store the tree-by-tree code, exceeding the available MCU flash capability. To reduce the model memory footprint and enable load monitoring at the edge, we chose to limit the number of trees. The configuration with 5 features and 100 trees provides a good trade-off achieving 95.15\% accuracy with 1.04\% drop compared to 96.19\% obtained using 71 features and 500 trees. Figures \ref{fig:RF}a and \ref{fig:RF}b shows the memory footprint and worst-case branch number for the two configurations. \\
As shown in Table~\ref{tab:rf_singleappliance}, reducing the 71-dimensional feature vector to 5 components leads to almost 2\% precision and recall drop. Thus, absolute values remain high, around 92\%. Recall results denote the model’s ability to recognize almost all relevant loads without skipping load activations. On the other side, high precision means that only real activations are detected, with very low false positives. The extraction process requires 105~Kcycles for both configurations, but what distinguishes the configurations is the classification stage. The optimized configuration requiring 4.84~Kcyles leads to 5.45$\times$ speedup and 80.57\% Flash usage decrease. Regarding the overall framework, the speedup slows down to 1.2$\times$ since the extraction execution time is one order of magnitude larger, while the Flash decrease remains high (78.86\%). 
\begin{table*}[h]
    \centering
    \setlength\extrarowheight{1.5pt}
    \resizebox{\textwidth}{!}{
    \begin{tabular}{V{3} l V{3} c|c V{3} c|c V{3} c|c|c|c|c|c|c V{3}} 
        \hlineB{3}
        \multirow{3.5}{*}{\textbf{Feature Vector Dim}} & \multicolumn{2}{c V{3}}{\textbf{F. Extraction}} & \multicolumn{2}{c V{3}}{\textbf{RF}} & \multicolumn{7}{c V{3}}{\textbf{Overall Framework (F. Extraction + RF)}} \\
        \cline{2-12}
                    & \thead{Flash\\(kB)} & \thead{Cycles\\(K)} & \thead{Flash\\(kB)} & \thead{Cycles\\(K)} & \thead{Flash\\(kB)} & \thead{Flash\\Red. (\%)} & \thead{Cycles\\(K)} & Speedup & \thead{Accuracy\\(\%)} & \thead{Precision\\(\%)} & \thead{Recall\\(\%)} \\ 
        \hlineB{3}
        71 (500 trees)	       & 14.1 & 105 & 649.4 & 26.4 & 663.5 & /    & 131.4 & /     & 96.19 & 93.99 & 93.99\\
        \hline
        5 (100 trees)          & 14.1 & 105 & 126.2 & 4.84 & 140.3 & 78.86 & 109.84 & 1.2x & 95.15 & 92.04 & 91.83\\
        \hlineB{3}
    \end{tabular}
    }
    \caption{\textbf{RF-based Single-Appliance \#1 Load Detection Performance On ARM Cortex-M4}}
    \label{tab:rf_singleappliance}
\end{table*}

\subsubsection{SVM in Single-Appliance Scenario \#2}
\begin{figure}[h]
    \centering
    \captionsetup{justification=centering}
    \includegraphics[width=0.7\linewidth]{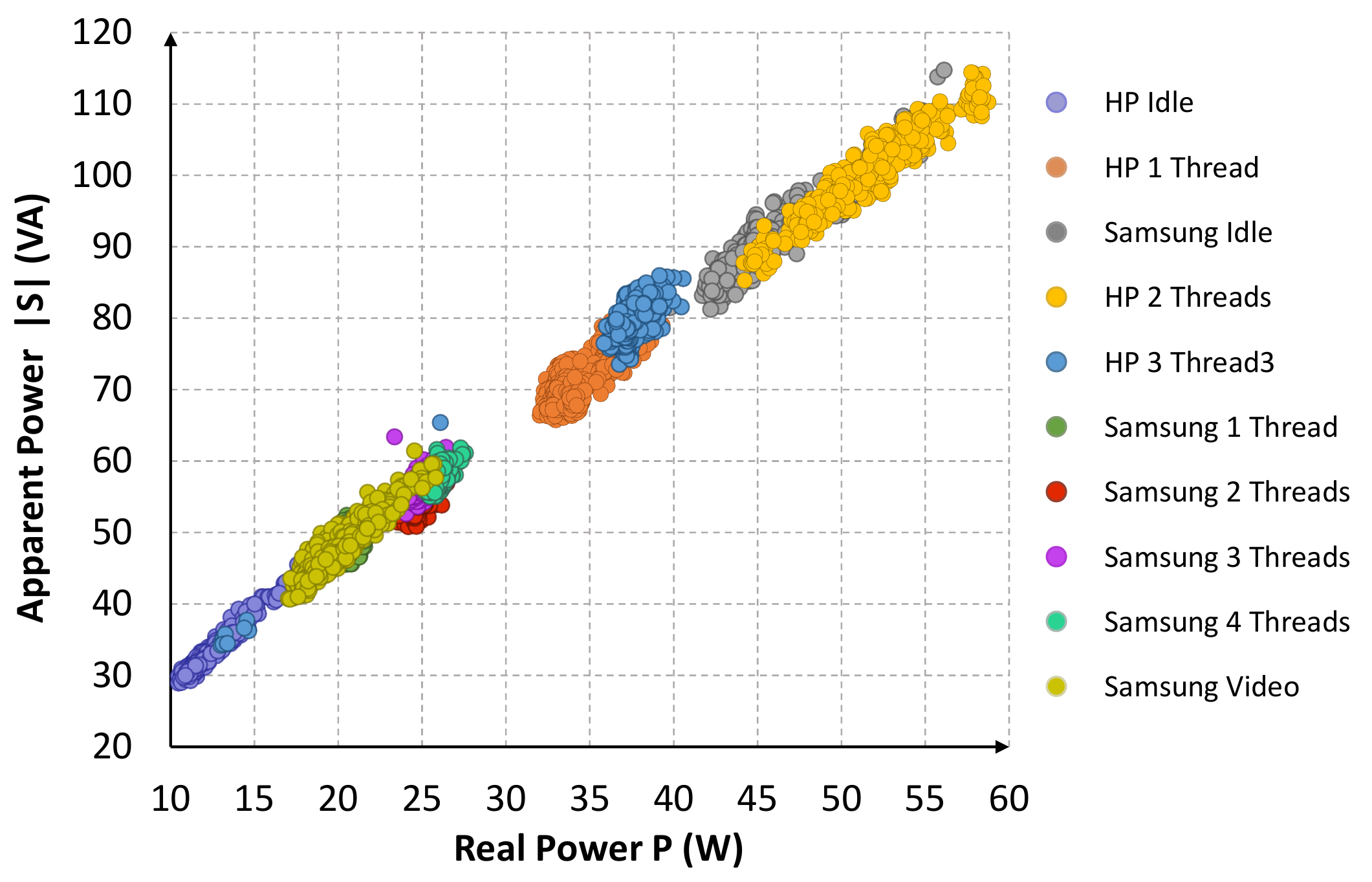}
    \caption{\textbf{Single-Appliance \#2 Dataset Instances}}
    \label{fig:classDistribution}
\end{figure}

\begin{figure}[ht]
    \centering
    \captionsetup{justification=centering}
    \includegraphics[width=0.6\linewidth]{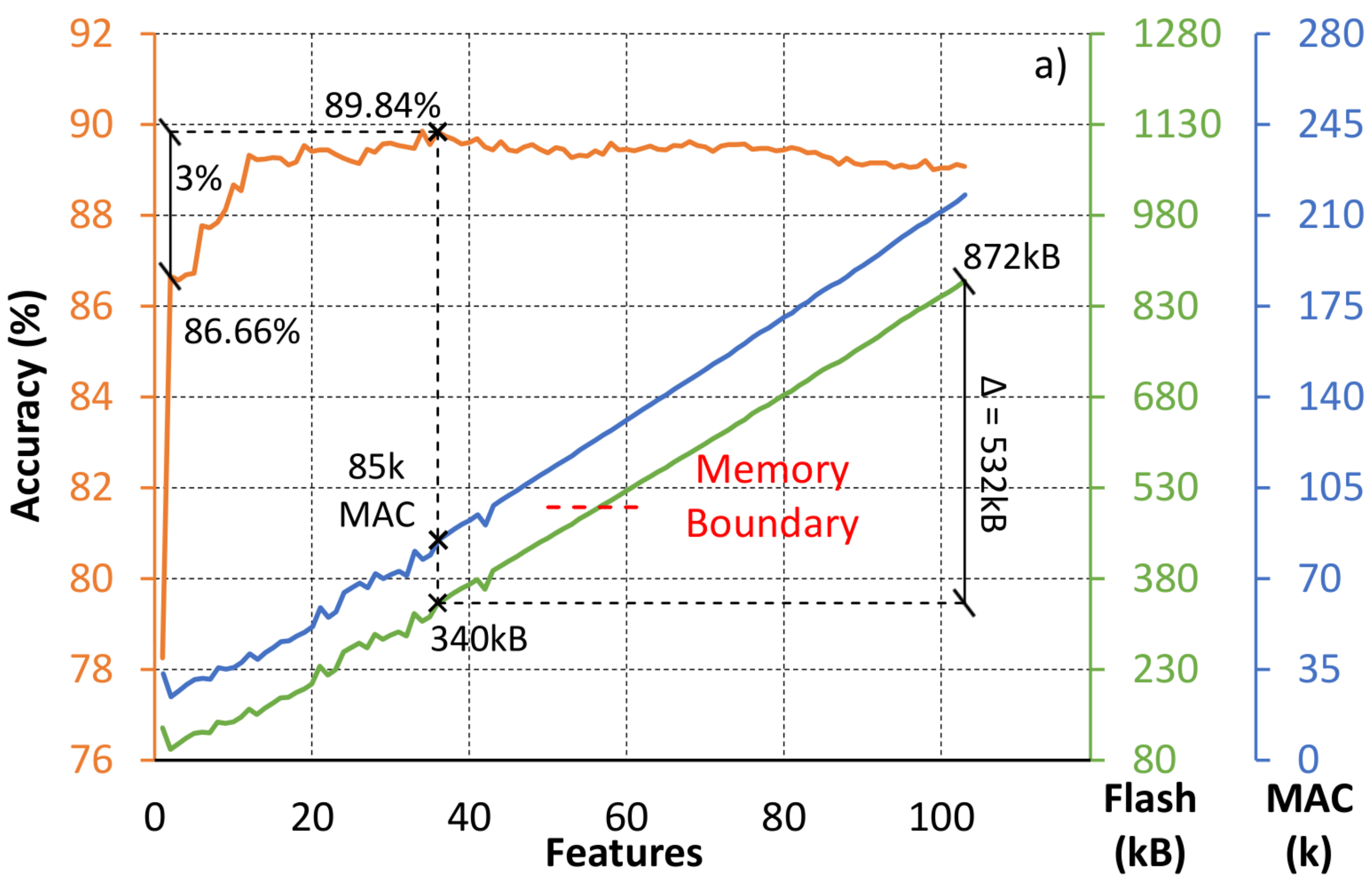}
    \caption{\textbf{SVM Accuracy-Flash-MACs Trade-Offs on Single-Appliance \#2 Scenario}}
    \label{fig:SVM1}
\end{figure}

\begin{figure}[ht]
    \centering
    \captionsetup{justification=centering}
    \includegraphics[width=0.6\linewidth]{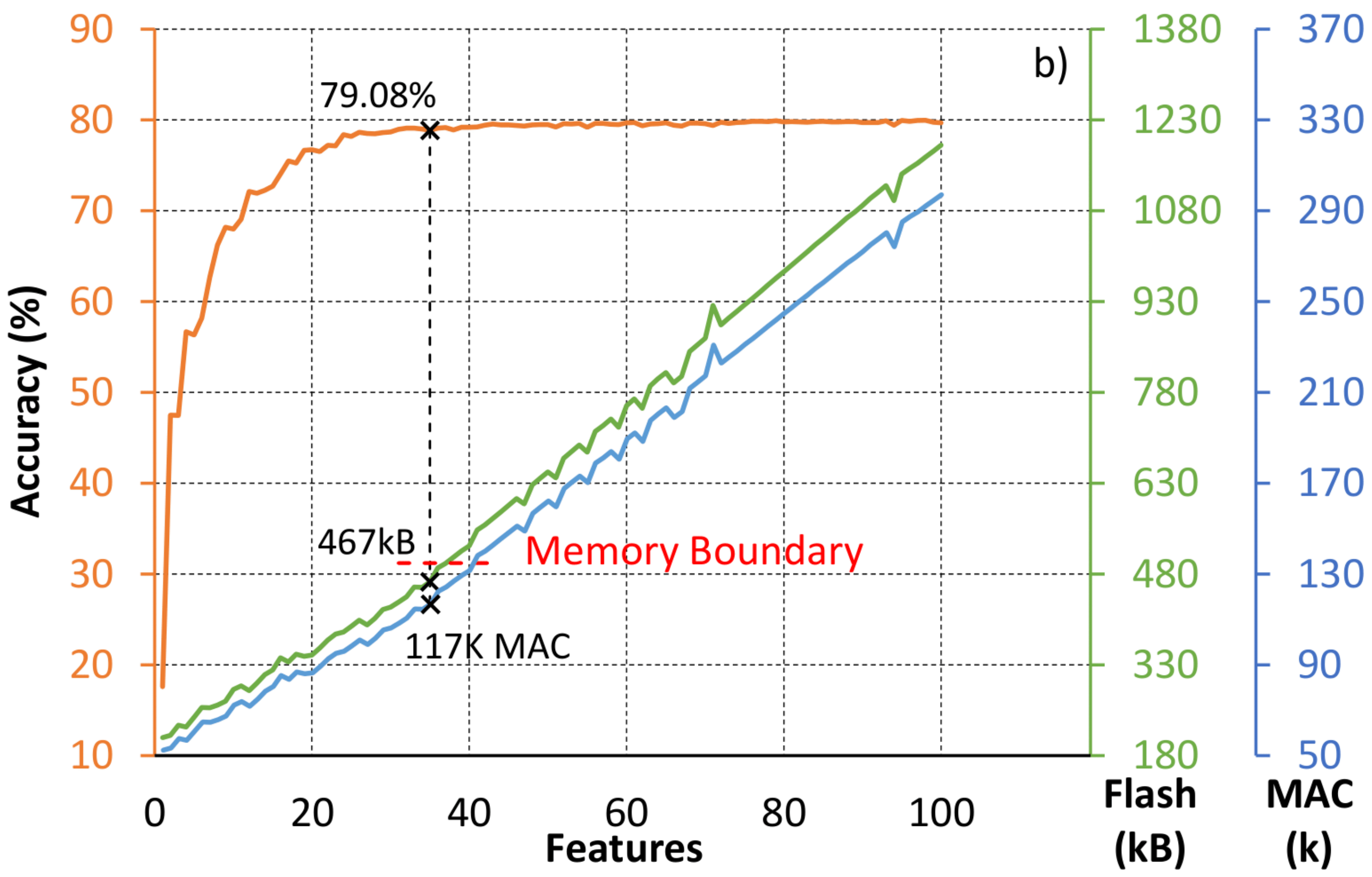}
    \caption{\textbf{SVM Accuracy-Flash-MACs Trade-Offs on Single-Appliance \#2 Scenario\\with only frequency-domain features}}
    \label{fig:SVM2}
\end{figure}

In this section, we analyze the trade-off involved in the application of SVM to a different Single-Appliance load monitoring scenario. The scenario reflects a more challenging context where the electric loads have similar time-domain feature distributions, making the recognition process harder. The deployed dataset consists of HP and Samsung laptops, which varies the thread count and the running task resulting in 10 classes. To highlight the challenge in recognizing these loads, in Figure \ref{fig:classDistribution}, we represented the dataset in a $P$ vs. $|S|$ graph, which are the most significant features according to the MDA analysis. Instances are colored according to the classes. From the plot, we observe that clusters partially overlap (e.g., HP 1 Thread/Samsung Idle and HP 2 Threads/HP 3 Threads), and in some cases (e.g., Samsung Video/Samsung 1 Thread and Samsung 3 Threads/Samsung 4 Threads) distinguishing them relying only on $P$ and $|S|$ is not possible. For that purpose, the additional contribution given by frequency-domain features is required.
\\
In Figure~\ref{fig:SVM1}, we show testing accuracy, MAC, and Flash usage trends when adding one MDA-ordered feature at a time in the feature space. On the left-most side of the plot, deploying only 1 feature demands singular high resources due to the dataset’s hard linear separability with low-dimensional feature spaces. Thereby, a large number of SVs is mandatory to maximize the margin around the separating hyperplane. Increasing the space dimension to 2 features improves data separability. The SVM needs fewer SVs to find the optimal hyperplane, leading to a minimum resource requirement. Expanding the space dimension further has a reduced impact on the number of SVs. Consequently, computing and memory efforts grow almost linearly with the increase of dimensionality.
\\
Using only time-based features ($P$ and $|S|$, first two left-most side points), the model achieves 86.66\% accuracy. Adding frequency-based features, we supply extra information useful to improve device recognition. The accuracy grows up to 89.84\% at 36 features, obtaining an incremental gain for each added feature till reaching a plateau. Using more features causes SVM overfitting leading to a slight accuracy degradation, reaching 89.07\% accuracy in the full feature case. Moreover, a full feature vector implementation would require almost 103-dimensional 1.95K floating-point SVs and 17.55K Dual Coefficients, meaning almost 872~kB Flash storage and 218K~MAC to run the inference. As shown in Table \ref{tab:Single_Appliance_Complex_2_SVM}, decreasing feature vector dimensionality from 103 to 36 results in a slight precision and recall increase ($\sim$0.8\%) due to SVM overfitting when using 103 features. However, in absolute value, results are lower ($\sim$2\%) than in the previous scenario. This result can be explained by the fact that the Single-Appliance \#2 scenario consists of appliances characterized by more similar time-domain feature distributions, complicating the identification task. This overall leads, in terms of results, to a higher fraction of both false positives and negative. The new optimized feature space  requires about 344~kB Flash storage and 795~Kcycles for the processing stage, while the extraction stage demands 105~Kcycles and few kBs. As a result, the overall system achieves 2.41$\times$ speed up and a Flash decrease of 59.63\% with respect to the full feature vector implementation. Optimizing the run-time with loop unrolling and improving the allocation of registers by placing accumulation variables into local registers, we achieve an execution time of 562~Kcycles, leading to a 3.25$\times$ overall speedup.
\begin{table*}[h]
    \centering
    \setlength\extrarowheight{1.5pt}
    \resizebox{\textwidth}{!}{
    \begin{tabular}{V{3} c V{3} c|c V{3} c|c V{3} c|c|c|c|c|c|c V{3}} 
        \hlineB{3}
        \multirow{3.5}{*}{\textbf{Feature Vector Dim}} & \multicolumn{2}{c V{3}}{\textbf{F. Extraction}} & \multicolumn{2}{c V{3}}{\textbf{SVM}} & \multicolumn{7}{c V{3}}{\textbf{Overall Framework (F. Extraction + SVM)}} \\
        \cline{2-12}
                    & \thead{Flash\\(kB)} & \thead{Cycles\\(K)} & \thead{Flash\\(kB)} & \thead{Cycles\\(K)} & \thead{Flash\\(kB)} & \thead{Flash\\Red. (\%)} & \thead{Cycles\\(K)} & Speedup & \thead{Accuracy\\(\%)} & \thead{Precision\\(\%)} & \thead{Recall\\(\%)} \\ 
        \hlineB{3}
        103	                & 14.1 & 105 & 871.9 & 2065 & 886 & /  & 2170 & /    & 89.08 & 89.06 & 89.16 \\
        \hline
        36                  & 14.1 & 105 & 343.55 & 795  & 357.65 & 59.63 & 900  & 2.41x & 89.84 & 89.86 & 89.90 \\
        \hline
        36 + Optimization   & 14.1 & 105 & 343.61 & 562  & 357.71 & 59.63 & 667  & 3.25x & 89.84 & 89.86 & 89.90 \\
        \hlineB{3}
    \end{tabular}
    }
    \caption{\textbf{SVM-based Single-Appliance \#2 Load Detection Performance on ARM Cortex-M4}}

    \label{tab:Single_Appliance_Complex_2_SVM}
\end{table*}
\\
To further test frequency-domain features ability to enhance load recognition in complex scenarios, we also trained the SVM model leaving out time-domain features (P, |S| and Q). The accuracy trend in Figure~\ref{fig:SVM2} reveals that identifying challenging devices is still possible, with approximately 80\% reached near 35 features. Then the model gets to a plateau with no valuable enhancement. Deploying 35 features demands 467~kB Flash storage and 117K~MAC to run the SVM inference, meaning an edge implementation is feasible. Moreover, since frequency-domain features rely only on current harmonics, its deployment would allow removing voltage sensors from the smart meter resulting in a substantial bill of materials cost reduction. 

\subsubsection{MLP in Multi-Appliance Scenario}
\begin{figure}[ht]
    \centering
    \includegraphics[width=0.6\linewidth]{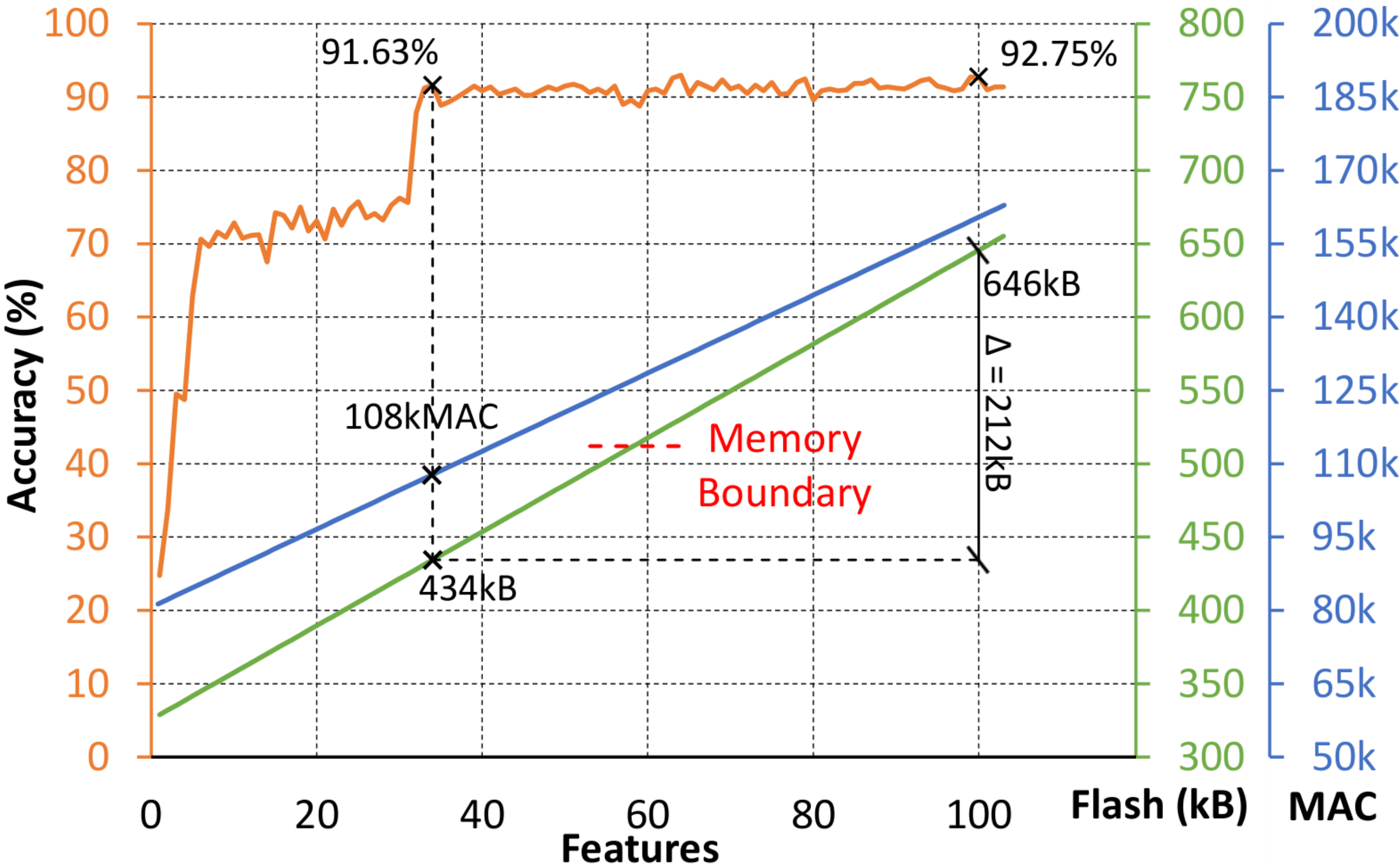}
    \caption{\textbf{MLP-based Multi-Appliance load disaggregation performance}}
    \label{fig:MLP}
\end{figure}

The Multi-Appliance Scenario has been developed to enable load disaggregation in a more-realistic context, where appliances can overlap. We deployed a dataset consisting of domestic loads totaling 5 classes: fan at minimum speed, electric coffee machine, light bulb, monitor, and power bank. The Grid Search led to 2-Layers MLP with 800 and 100 neurons as the most performing architecture. In Figure~\ref{fig:MLP}, we represent testing accuracy, MAC, and Flash usage trends adding one MDA-ranked feature at a time in the feature space. Deploying low-dimensional feature spaces (left-most side of the plot) requires low resource usage but it is not accurate enough. Increasing the feature space dimension makes growing memory and computational costs linearly, while MLP accuracy boosts up to a plateau at 34 features, resulting in no significant accuracy improvement beyond that. When adding the 32nd feature (Reactive Power - $Q$), the accuracy has a significant increase passing from 75.63\% to 91.25\%. The jump is due to the lack of correlation information between features of the MDA analysis.
\begin{table*}[h]
    \centering
    \setlength\extrarowheight{1.5pt}
    \resizebox{\textwidth}{!}{
    \begin{tabular}{V{3} l V{3} c|c V{3} c|c V{3} c|c|c|c|c|c|c V{3}} 
        \hlineB{3}
        \multirow{3.5}{*}{\textbf{Feature Vector Dim}} & \multicolumn{2}{c V{3}}{\textbf{F. Extraction}} & \multicolumn{2}{c V{3}}{\textbf{MLP}} & \multicolumn{7}{c V{3}}{\textbf{Overall Framework (F. Extraction + MLP)}} \\
        \cline{2-12}
                    & \thead{Flash\\(kB)} & \thead{Cycles\\(K)} & \thead{Flash\\(kB)} & \thead{Cycles\\(K)} & \thead{Flash\\(kB)} & \thead{Flash\\Red. (\%)} & \thead{Cycles\\(K)} & Speedup & \thead{Accuracy\\(\%)} & \thead{Precision\\(\%)} & \thead{Recall\\(\%)}\\
        \hlineB{3}
        100	                & 14.1 & 105 & 645.62 & 1588  & 659.72 & /     & 1693  & /     & 92.75 & 91.69 & 91.82\\
        \hline
        34                  & 14.1 & 105 & 432.7 & 1081  & 446.8 & 32.27 & 1186  & 1.43x & 91.63 & 90.02 & 90.49\\
        \hline
        34 + Optimization   & 14.1 & 105 & 434.1 & 861.7 & 448.2 & 32.06 & 966.7 & 1.75x & 91.63 & 90.02 & 90.49\\

        \hlineB{3}
    \end{tabular}
    }
    \caption{\textbf{MLP-based Multi-Appliance Load Detection Performance On ARM Cortex-M4}}
    \label{tab:MLP}
\end{table*}
\\
As shown in Table~\ref{tab:MLP}, reaching the top accuracy point (92.75\%) demands almost a full feature vector deployment (100 features). However, the model requires 646~kB Flash memory, which is well above the MCU capability. Limiting the feature vector to 34 components, MLP achieves 91.63\% accuracy with a 1.12\% drop regarding the top accuracy point. Decreasing the feature vector dimensionality increases false positives and negatives, leading to precision and recall drop ($\sim$1.5\%). Compared to the Single-Appliance \#1 scenario, the Multi-Appliance scenario consists of switching appliances that can overlap. As a result, overall accuracy, precision, and recall are lower than scenario \#1 but still acceptable, highlighting the capability of distinguishing loads among each other. Moreover, the optimized feature vector decreases the MLP size by 32.27\%, and with 1.082~Mcycles to run an inference speedups the execution time of 1.47$\times$. Adopting run-time optimizations, such as loop unrolling and registers allocation improvement, reduces the execution time to 861.7~Kcycles leading to a 1.84$\times$ speed-up. \\
From the experimental evaluation of other dimensionality reduction methods, we conclude by noting that in the case of deployments specifically tailored to multi-appliance scenarios, the PCA achieves better results in terms of recall (94.17\%) and precision (92.58\%), and comparable accuracy results.

\subsection{Overall Framework Characterization}
\label{sec:overall}
\begin{figure}[h]
    \centering
    \includegraphics[width=1\linewidth]{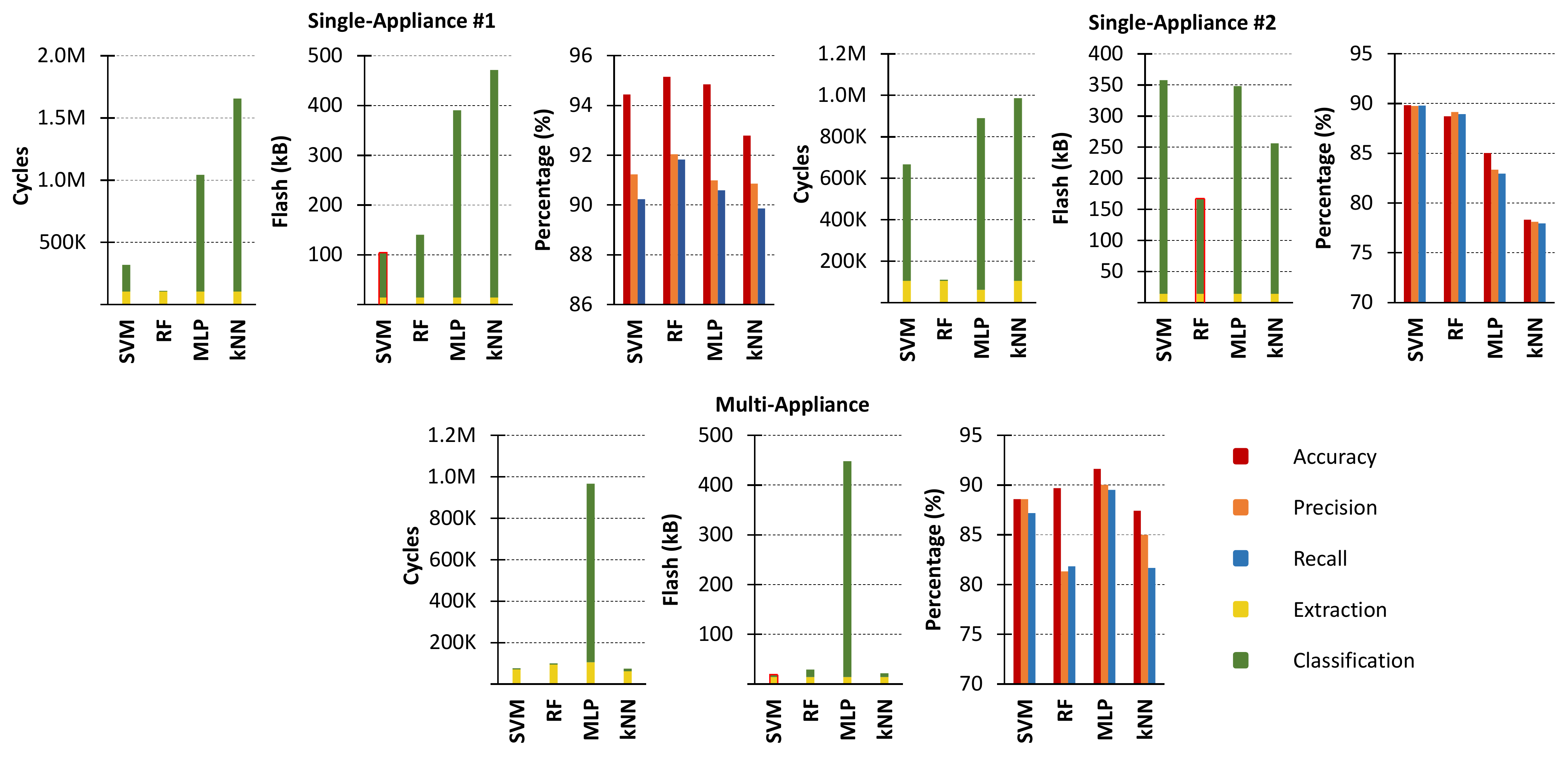}
    \caption{\textbf{Overall Framework Comparison}}
    \label{fig:FinalComparison}
\end{figure}

By combining the MDA analysis with the memory-performance-accuracy trade-off evaluation, we report the NILM framework characterization in each scenario. In Figure \ref{fig:FinalComparison}, we show the computing and memory cost required by NILM algorithms with optimized feature spaces, highlighting the frameworks featuring the smallest Flash footprint with red borders. Finally, we summarize in Table \ref{tab:Results} advantages and disadvantages of using each algorithm. We reported the resources and the metric from which an edge-based NILM implementation depends more: memory occupation, execution time, and accuracy. The '$+$' symbol represents a trend highly fittable in resource-constrained MCUs (low memory consumption, low latency, and high accuracy). In contrast, the '$-$' symbol highlights a tendency that is highly likely to make unfeasible the adoption of the model on-the-edge with different scenarios and setups.

\section{Discussion}
The obtained results demonstrate the feasibility of edge-based NILM systems. The methodology used allows reducing feature dimensionality without undermining load monitoring accuracy in different scenarios.\\
In the Single-Appliance \#1 context, we show that reducing a 103-dimensional feature vector to 5 components results in a RF model with a modest accuracy drop but enables 80.56\% Flash usage reduction and 5.45$\times$ speedup. We demonstrate that the use of frequency-domain features leads to an additional 3\% contribution to load recognition accuracy when time-domain components present similar distributions. Moreover, we explored using only frequency-domain features with almost 80\% accuracy, leading to a substantial frontend cost reduction, as voltage sensors can be removed if the accuracy loss is deemed to be acceptable.\\
When multiple loads are active simultaneously, we prove that, by reducing the features to a 34-dimensional vector, a 2-Layer MLP model reaches 91.63\% accuracy requiring 448kB Flash memory and almost 862Kcycles, corresponding to a small execution time of 10.26msec.\\
Our research provides clear evidence that on-the-edge load monitoring is possible, as we can reduce model complexity to fit low-cost MCU-based meters memory and computational capabilities. Along this path, it is possible to foresee the application of edge NILM to innovative services such as Home Energy Management (HEM) and Anomaly Detection (AD), entailing more complex multi-appliance scenarios with additional load types. However, this would require a larger amount of training data to feed the training pipeline and complex automatic data annotation systems to address unknown novel appliances. Moreover, the feature extraction on 100ms-long time windows might be insufficient for identifying intermediate power states of complex FSM loads. All these are directions of future work.
\par

\begin{table}[h]
    \centering
    \setlength\extrarowheight{1.5pt}
    \begin{tabular}{ c|c|c|c}
        \hlineB{3}
                & \textbf{Latency} & \textbf{Memory} & \textbf{Accuracy} \\ 
        \hlineB{3}
        SVM     & $+$ & $-$ & $+$ \\
        \hline
        RF      & $+$ & $+$ & $+$ \\        
        \hline
        MLP     & $-$ & $-$ & $+$\\
        \hline
        kNN     & $-$ & $-$ & $-$\\
        \hlineB{3}
    \end{tabular}
    \vspace{0.2cm}
    \caption{\textbf{Algorithm Advantages and Disadvantages}\\}
    \label{tab:Results}
\end{table}

\section{Conclusion}
The paper presents a novel strategy to lighten NILM framework complexity, thus enabling moving intelligence to the edge. We developed the study using a flexible and low-power \emph{Smart Measurement Node} which features an advanced analog front-end for dual-channel voltage and current 14-bit acquisition and 1.5Msps sampling rate.\\
After highlighting that model complexity and feature vector size are highly correlated, we performed a MDA analysis to reduce the feature space dimensionality without endangering information content. Results reveal the most important features among different scenarios, thus enabling lowering the dimension with minimal accuracy loss. Comparing memory, latency, and accuracy of NILM algorithms, we brought tangible benefits in lightening feature extraction and classification workloads by deploying reduced feature spaces and optimized run-time. We compare 4 supervised learning techniques available in the literature on 3 different load classification scenarios. The study demonstrates that a feature vector reduction to lowering the computing effort and memory-footprint is achievable without undermining NILM accuracy. \\
Future work will improve the system execution time using the second ultra-low-power multi-core available on the meter. Furthermore, we will investigate in more depth the capacity of current harmonics to disaggregate loads while analyzing how different time-window lengths affect the framework accuracy and extraction cost. 

\end{document}